\documentclass[runningheads]{llncs}
\usepackage[T1]{fontenc}
\usepackage{graphicx}
\usepackage{subcaption}
\usepackage{booktabs}
\usepackage[misc]{ifsym}
\usepackage{subfiles} 
\newcommand{\corr}{(\Letter)}

\newcommand{\floor}[1]{\left\lfloor #1 \right\rfloor}
\newcommand{\ceil}[1]{\left\lceil #1 \right\rceil}
\usepackage{mwe}

\begin{document}

\title{An Empirical Evaluation of Factors Affecting SHAP Explanation of Time Series Classification}

\titlerunning{Factors Affecting SHAP Explanation of TSC }

\author{Nikos Papadeas \inst{1}\textsuperscript{*} \and
Davide Italo Serramazza \corr \inst{2}\textsuperscript{*} \and 
Zahraa Abdallah\inst{1} \and
Georgiana Ifrim\inst{2}
}

\authorrunning{D. Serramazza, N. Papadeas et al}

\institute{School of Engineering Mathematics and Technology, University of Bristol, UK 
\and
School of Computer Science, University College Dublin, Ireland 
}

\maketitle 
\textsuperscript{*}These authors contributed equally to this work.

\begin{abstract}
Explainable AI (XAI) has become an increasingly important topic for understanding and attributing the predictions made by complex Time Series Classification (TSC) models. 
Among attribution methods, SHapley Additive exPlanations ({SHAP}) is widely regarded as an excellent attribution method, but its computational complexity, which scales exponentially with the number of features, limits its practicality for long time series. 
To address this, recent studies have shown that aggregating features via segmentation, to compute a single attribution value for a group of consecutive time points, drastically reduces SHAP running time. However, the choice of the optimal segmentation strategy remains an open question. 
In this work, we investigated eight different Time Series Segmentation algorithms to understand how segment compositions affect the explanation quality. We evaluate these approaches using two established XAI evaluation methodologies: InterpretTime and AUC Difference.
Through experiments on both Multivariate (MTS) and Univariate Time Series (UTS), we find that the number of segments has a greater impact on explanation quality than the specific segmentation method. Notably, equal-length segmentation consistently outperforms most of the custom time series segmentation algorithms. Furthermore, we introduce a novel attribution normalisation technique that weights segments by their length and we show that it consistently improves attribution quality. 

\end{abstract}

\section{Introduction}
\label{sec:intro}

Time series (TS) data is a common data type in domains including finance, healthcare, sports analytics, and market predictions.
The increasing complexity of Machine Learning (ML) models led to the emergence of Explainable Artificial Intelligence (XAI), a field aiming to make AI decisions interpretable, to avoid the risks of relying on faulty predictions in high-stakes domains.

In the context of time series XAI, several studies have adapted SHAP \cite{lundberg2017unified}, arguably one of the most popular XAI methods, to the time series domain. In addition to time series-specific challenges, such as temporal and channel dependencies or locality, SHAP also poses computational challenges as it scales exponentially with the number of features. Past works in the TS domain mainly used \textit{segmentation} or \textit{chunking}, i.e., explaining a group of consecutive time points together, assigning them the same attribution, thus decreasing the number of features to be explained. This technique has proven to be effective in lowering the computational demand by a few orders of magnitude \cite{seramazza24}, but to the best of our knowledge, no systematic study has investigated the impact of different segmentation strategies on the quality of the resulting explanation.

This work investigates eight different segmentation methods, including the popular and simple \textit{equal-length segmentation}, using a varied range of Univariate and Multivariate Time Series Classification (UTSC/MTSC) problems and four different classifiers. 
Additionally, we also explored the impact of the background set used by SHAP, i.e., the instances to be used when simulating data missingness: we compared the simple \textit{zero} background, the default choice for some important frameworks such as Captum \cite{kokhlikyan2020captum} with the \textit{average background}, where missing values are replaced with the average value for each observation in the training set. 

Moreover, SHAP values for segment-based explanations are computed over time segments rather than individual timepoints, but evaluations require single-timepoint attributions. In order to maintain segment-level additivity of individual timepoint attributions, we introduce a normalisation method based on segment length.  

Finally, while various approaches have been proposed for the quantitative or qualitative evaluation of XAI methods, in this work, we focus on two quantitative approaches designed for evaluating time series XAI methods: \textit{InterpretTime} \cite{turbe2023evaluation} and \textit{AUC Difference} \cite{aucd,rise}.

Our main contributions in this  paper are:
\begin{itemize}
    \item We evaluate eight segmentation models using a wide variety of datasets and ML models. Our experiments clearly show that the best choice is the simple \textit{equal segmentation} with some notable exceptions, which have been further analysed.
    \item We showed that the average and the zero background performances are very often at the same level; however, the average background performs better in some cases. 
    \item  We proposed a \textit{normalisation} technique which weights every segment attribution according to its length. Our experiments show that in the vast majority of cases, this approach improves the XAI evaluation scores.
    \item To help further research in this area, we publicly release our code and data.\footnote{\url{https://github.com/davide-serramazza/segment\_SHAP}}
\end{itemize}
\section{Related Work}
\label{sec:relwork}

We discuss in this section relevant methods for time series classification, segmentation, explanation, and the evaluation of XAI methods.

\subsection{Time Series Classification}

\textbf{QUANT} \cite{dempster2024quant} is a TSC algorithm that relies exclusively on quantile-based features. The method extracts features using fixed dyadic intervals from four distinct representations of the input TS, namely the raw TS, its Fourier transform, and its first and second order differences. These 4 different feature representations are then concatenated before being fed into an \textit{Extra Trees classifier} \cite{extratrees}. 

\textbf{MiniRocket}  \cite{miniRocket} is part of the popular Rocket TSC algorithm family: the main idea is to extract features by sliding a large number of convolutional kernels over the TS. The main feature is \textit{PPV (Proportion of Positive Values)}, the proportion of positive values as output of the kernel, used as a proxy to capture the ones best matching with the current TS. Once the features are extracted, a linear model is used as a classifier, in our case, a Ridge Classifier.

\textbf{ResNet} \cite{he2016deep} is a \textit{Convolutional Neural Network (CNN)} that has been proposed for image classification and then applied to TSC. The main model innovation is the \textit{residual connection}, i.e., skipping one or more layers to allow the network to learn a deeper representation, reducing the vanishing gradient problem. Notably, it has been one of the first deep learning models to be successfully applied to TSC without any architectural changes compared to the original one.

\textbf{Random Forest} is an ensemble learning method that constructs multiple decision trees during training and outputs the mode of their prediction. Despite being a \textit{tabular data} classifier, it yields good performance when applied to TSC \cite{bhaskar23}; however, this model can only handle $1D$ data, thus in the case of MTS, channels are concatenated to create a $1D$ UTS.

\subsection{Time Series Segmentation}

\textbf{Equal-length segmentation:} The first segmentation we evaluated is the equal length one: a TS of length $L$ is divided into $n$ different segments of length $ [\floor{\frac{L}{n}}, \ceil{\frac{L}{n}}]$.

\textbf{Classification Score Profile (ClaSP)} 
\cite{ermshaus2023clasp} evaluates each possible split by training a binary classifier to identify subsequences from both sides. After scoring each possible point, it applies peak detection to select the actual change points (CPs). ClaSP works dimension-wise; thus we apply ClaSP independently for each dimension when using MTS data, possibly finding an uneven number of segments for each channel.

\textbf{Information Gain Segmentation(IG)} \cite{sadri2017information} aims to maximise the amount of entropy lost due to a possible segmentation: for each of the change points to be detected, it uses a top-down search to evaluate the information gain resulting from the possible next changing point.

\textbf{Greedy Gaussian Segmentation (GG)} \cite{hallac2019greedy} finds the segmentation by interpreting the data as coming from a Gaussian distribution. Precisely, it finds an approximate solution to the maximum log-likelihood of a segmented Gaussian model having the specified number of changing points.

\textbf{NNSegment} comes from \textit{LimeSegment} \cite{limeSegment}, it detects change points in TS by assuming that each observation belongs to either a repeating motif or an anomaly. After having split the TS into sub-windows $w_i$, it assigns each $w_i$ its nearest neighbours $w_j$. If adjacent windows $w_i$ and $w_j$ have different nearest neighbours, a change point is flagged either at the window beginning or at its end. Similarly to ClaSP, NNSegment is applied dimension-wise.

\textbf{Binary Segmentation (Binseg)} \cite{sen1975tests} is a greedy algorithm that uses a recurrent approach. Starting from the entire series, it uses a cost function to detect the most significant CP within the current segment. After splitting, the operation is repeated on the two resulting sub-series until the required number of changing points is detected.

\textbf{Bottom-up segmentation (BottomUp)} \cite{killick2012optimal} starts with an over-split of the current TS in regular, fixed-length segments. Contiguous sub-samples are then merged according to a cost function until the target number of changing points is reached. 

\textbf{Kernel change point detection (KernelCPD)} \cite{celisse2018new} maps the TS into a Hilbert space $\mathcal{H}$ using a kernel $\mathcal{K} $. Change points are detected by finding the mean-shifts in the mapped signals in $\mathcal{H}$.

\subsection{Evaluation of XAI for Time Series Classification}

\textbf{InterpretTime} was proposed in \cite{turbe2023evaluation}, and the main idea is to perturb samples in the test set based on the importance ranking generated by the evaluated attribution method. Specifically, TS observations are progressively corrupted by injecting values sampled from a distribution (default is the Normal distribution), following the importance rank specified by the attribution values. 
Specifically, the method involves:

\begin{itemize}
    \item Establishing perturbation thresholds $k = [0.05, 0.15, ..., 0.95, 1.0]$
    
    \item   Generating two modified versions for each $k$, $\bar{X}^{top}$ and $\bar{X}^{bottom}$ replacing respectively the top $k$ and the bottom $(1-k)$ quantiles of the original TS values with values sampled from the specified distribution. Top $k$ and bottom $(1-k)$ quantiles are identified only considering the {positive attribution values} of the explanation method to be scored.

    \item Computing normalised score differences: $\bar{S} = \frac{S(X) - S(\bar{X})}{S(X)}$, where $S$ represents the probability for the initial predicted category.
\end{itemize}

 Based on $\bar{S}$, two metrics are defined, namely \textbf{AUCSE}, which quantifies how effectively the method identifies the most important TS features, and \textbf{F-score}, measuring the balance between detecting high-importance and low-importance features. InterpretTime was shown to be effective in ranking attribution methods and to have a good correlation between the quantitative ranking and the qualitative method evaluation. 
 The methodology was further shown to benefit from using multiple perturbations, beyond the default Gaussian: in this work, we used this extended variant of InterpretTime \cite{seramazza24}.

 \textbf{AUC Difference (AUCD)} \cite{aucd,rise}  assesses attribution quality by analysing how perturbations of time-series observations, guided by importance rankings, influence model predictions. This evaluation relies on two complementary perturbation approaches: deletion and insertion. Both approaches are based on ordering timepoints based on their saliency as defined by the attribution method that is being evaluated, and then replacing those timepoints and measuring the change in prediction. The replacement values are derived from the \textit{opposite class}, defined as the class representative (the mean of each sample in a class) that has the lowest prediction for the explained sample's predicted class.

During deletion, the method iteratively replaces the most salient timesteps of the explained sample with those from the opposite class. As these critical timesteps are removed, the model's prediction confidence diminishes. The area under this deletion curve, known as AUDC, serves as a key measure, where superior explanations produce smaller AUDC values, reflecting steeper declines in confidence.

The insertion process works inversely by starting with the opposite class sample and progressively replacing its values with the original sample based on the same ordering as deletion. This reconstruction allows the model's confidence to rebound, with the area under the insertion curve, or AUIC, quantifying the recovery speed. Higher AUIC values indicate more effective attribution methods that accurately pinpoint influential timesteps.

By combining these measures, the AUCD measure emerges as the arithmetic difference between AUIC and AUDC. Optimal explanations approach a value of 1.0, demonstrating both strong confidence recovery during insertion and rapid confidence loss during deletion.

The most notable difference between the two evaluation measures is that AUCD ranks features based on their absolute attribution value, while InterpretTime discards non-positive attribution values, meaning that they are not considered during the perturbation process. This has implications for segments with negative attributions, where the perturbed segments which have positive attribution might not be sufficient to change the prediction significantly.  \\
The other key difference between the two evaluation measures lies in the substitution values used when perturbing the time series: while InterpretTime uses values sampled from the specified distribution, AUCD selects values from an instance of the opposite class.

\section{Attribution Normalisation}
For segment-based explanations, Shapley values are computed over segments of timepoints rather than individual timepoints. However, evaluation methods for time series require attributions at the granularity of single timepoints. To address this incompatibility, each timepoint within a segment can be assigned the Shapley value of its parent segment. Formally, if a segment \( S\) has an attribution \( \phi_S \), every timepoint \( f_i \in S \) is assigned:  

$$
\phi_{f_i} = \phi_S.
$$  

A key property of Shapley values in this setting is segment-level additivity, which requires that the sum of attributions across all timepoints within a segment equals the segment’s total Shapley value. This method violates segment-level additivity, as the sum of attributions within the segment becomes:  

$$
\sum_{f_i \in S} \phi_{f_i} = |S| \cdot \phi_S,
$$  

which only equals \( \phi_S \) when \( |S| = 1 \).  

Maintaining segment-level additivity is essential for ensuring that Shapley-based interpretations remain theoretically sound. If the sum of individual timepoint attributions within a segment deviates from the segment’s original Shapley value, the resulting explanations lose their cooperative game-theoretic justification. This inconsistency can distort timepoint importance rankings, misrepresent the true contribution of segments, and introduce artefacts when comparing attributions across different granularities (e.g., segments vs. individual timepoints). For this reason, we suggest to instead use normalisation that distributes the segment’s Shapley value uniformly among its constituent timepoints, assigning each \( f_i \in S \) an attribution of:  

$$
\phi_{f_i} = \frac{\phi_S}{|S|},
$$  

where \( |S| \) denotes the segment’s length.  

This proposed normalisation method preserves the additivity property by construction:

$$
\sum_{f_i \in S} \phi_{f_i} = \sum_{f_i \in S} \left( \frac{\phi_S}{|S|} \right) = \phi_S.
$$  

 The normalisation method, by preserving segment-level additivity, ensures that evaluations performed on individual timepoints remain faithful to the original segment-wise Shapley values, thereby supporting reliable and interpretable model explanations.

 In our experiments, we compare results in the presence and absence of normalisation to assess its effectiveness.
\section{Experiments}
In this work, we use the Shapley Value Sampling implementation for SHAP \cite{seramazza24}, which allows us to specify a segmentation of the time series, such that each segment is considered as a feature for SHAP.
In the rest of this section, we specify the implementation and hyperparameters for the segmentation methods, classifiers, and XAI evaluation methodologies used. Lastly, we introduce the two tested backgrounds and the datasets used in the experiments.

\subsection{Classifiers}

Every classifier was trained using the default hyperparameters; the implementations for MiniRocket and QUANT are from the aeon library \cite{aeon24jmlr}. For RandomForest, we used the scikit-learn implementation \cite{scikit-learn}, and for ResNet, we rely on the implementation from dCAM  \cite{boniol2022dcam}.

\subsection{Segmentation Methods}
We use the sktime \cite{loning2019sktime} implementations for IG and GG, Ruptures \cite{truong2020selective} implementations for  BinSeg, BottomUp, and KernelCPD and aeon for ClaSP. Finally, NNSegment was adapted from the LimeSegment repository \cite{limeSegment}, and equal-width segmentation uses the custom implementation from \cite{seramazza24}. \\
For hyperparameters, we only deviated from defaults for ClaSP’s \textit{period length}, which was set to $4$, because using the default value we found only one segment. Additionally, a kernel type for KernelCPD and the cost functions for BinSeg, BottomUp need to be specified: we set them respectively to \textit{linear kernel} and $L1$ \textit{cost function} to account for speed.

\subsection{XAI Evaluation Metholodologies}
For InterpretTime, we used the implementation in \cite{seramazza24}, which extends the original one of \cite{turbe2023evaluation} by complementing the use of Gaussian distribution to perturb TS, with multi-distribution perturbations (global Gaussian, global mean, local mean).  \\
These four different \textit{perturbation strategies} (hereafter referred to simply as perturbations) are a proxy to test the stability of AUCSE and F-score when the classifier training process used by the original work is altered to remove the \textit{Gaussian noise injection} (more detail can be found in \cite{turbe2023evaluation}). Preliminary results demonstrated that the noise injection step degrades the accuracy of non-deep learning classifiers.

For AUCD, we implemented our own version according to its original definition. The only tunable parameter in the method is the step size, which indicates the number of timepoints to be replaced at each iteration. We have set the step to $4 \%$  of the time series data.

\subsection{Background for SHAP}

SHAP uses a \textit{background set} to simulate data missingness: when feature(s) need to be analysed as inactive(s), they are substituted with corresponding values of the instances found in this set. \\
In the experiments, two different background sets $\mathcal{D}_{backg}$ are analysed, both are composed of a single instance, as opposed to the \textit{sampling approach} \cite{kokhlikyan2020captum}. Since the computational demand of SHAP linearly increases with the cardinality of the background set, this choice is made to account for speed.

\textbf{Zero}: a sample full of zeros, the default choice for some explanation frameworks such as \cite{kokhlikyan2020captum}, the one used for SHAP's implementation. Mathematically 

$$\mathcal{D}_{backg} = \{ \mathbf{0}_{d\times L} \}$$

where $d$ is the number of channels (in the univariate case is equal to $1$) and $L$ is the number of observations in the series.

\textbf{Average}: a TS representing the average of samples in the training set, mathematically 

$$\mathcal{D}_{backg}= \{ \frac{ \sum_{x_i \in \mathcal{D}_{train}} x_i }{ | \mathcal{D}_{train} |} \}$$.

\subsection{Datasets}
We use five diverse datasets in our experiments, covering both univariate and multivariate time series settings.

\noindent \textbf{Gunpoint (GNP)}\cite{ratanamahatana2005three} is a very well-known baseline dataset. It represents the $x$ coordinate position of a (simulated) gun when two different actors perform two different gestures (targets) \textit{draw} and \textit{point}. The TS are univariate, having 150 time points each. 

\noindent \textbf{UWAVE} \cite{liu2009uwave} data represent the $x,y,z$ accelerometer coordinates (the 3 dimensions) of hands while executing 8 different gestures (targets). The length of each MTS is 315. 

\noindent \textbf{EOG} is the concatenation of two different UTS datasets \cite{fang2018electrooculography} representing a horizontal and a vertical EOG signal. Four electrodes were placed on different sides of the left eye. The horizontal dimension is the difference between the left and right signals, while the vertical dimension is the difference between the upper and the lower signals. Each dimension has a length of 1250 time points, and the dataset has 12 different targets.

\noindent \textbf{KeplerLightCurves (KLC)} \cite{barbara2022classifying} data is obtained by light curves coming from NASA's Kepler mission; the 7 possible targets represent different classes of stars. Each instance has been labelled by experts working at the University of Sydney and interpolated such that every instance has a length of $4767$ time points.

\noindent \textbf{MilitaryPress (MP8)} \cite{dhariyal2023scalable}: 
TS represent $x,y$ coordinates of $25$ different body parts while executing the \textit{Military Press strength-and-conditioning} exercise. The dataset has 4 different targets representing an execution of the exercise, and each dimension has 161 time points. We used a \textit{domain-expert selection} containing 8 channels, i.e., the $y$ coordinates of both right and left positions of Elbows, Wrists, Shoulders, and Hips.

\section{Results}

\begin{figure}[t]
    \centering
    \includegraphics[width=\linewidth]{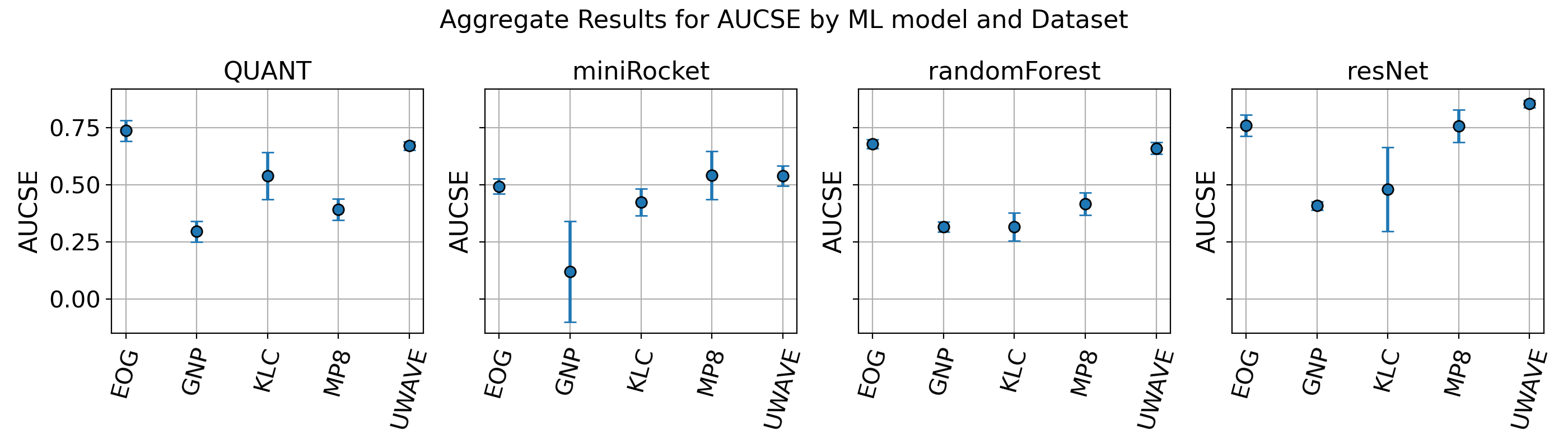} 
     \caption{Aggregated $AUCSE$ by dataset  and classifier. Error bars are standard deviation from the mean. AUCSE is computed based on the SHAP attributions for each classifier-dataset pair. }
    \label{fig:aggregated_AUCSE}

\end{figure}

For each dataset and machine learning (ML) model, we aggregate the results for each evaluation measure over the three factors of interest: backgrounds, perturbations, and segmentations. 
We calculate the mean and variance for each combination as shown in Figure \ref{fig:aggregated_AUCSE} for AUCSE (other evaluation measures have shown similar trends). This conditional variance analysis, where variance is examined with respect to fixed datasets and ML models, allows us to identify the cases where one or more of the three factors of interest are responsible for the variance in the results. Specifically, the results suggest that in most cases, there is little to no variance due to these factors, and conditioning on the dataset and ML model sufficiently determines the outcome of the evaluation measures, meaning that the impact of these factors is negligible. 

This very low variance also suggests that \textbf{equal-length segmentation is often the best choice}: some of the evaluated segmentation methods come with a high computational cost, which does not justify their marginal improvement, where it even exists, over the cost-free equal segmentation.

However, there are notable exceptions, where there is significant variance in specific combinations of data and classifier, such as MiniRocket on GNP and MP8 or ResNet on KLC. In these cases, one or more of the remaining factors (background or segmentation) appear to significantly influence explanation quality. Therefore, we take a deeper look into these specific cases to find out when the choice of background or segmentation does influence the quality of explanations according to each evaluation measure.

\subsection{Background and Segmentation Analysis}
\subsubsection{Segmentations on KeplerLightCurves.}

\begin{figure} [t]
    \centering
    \includegraphics[width=0.7\linewidth]{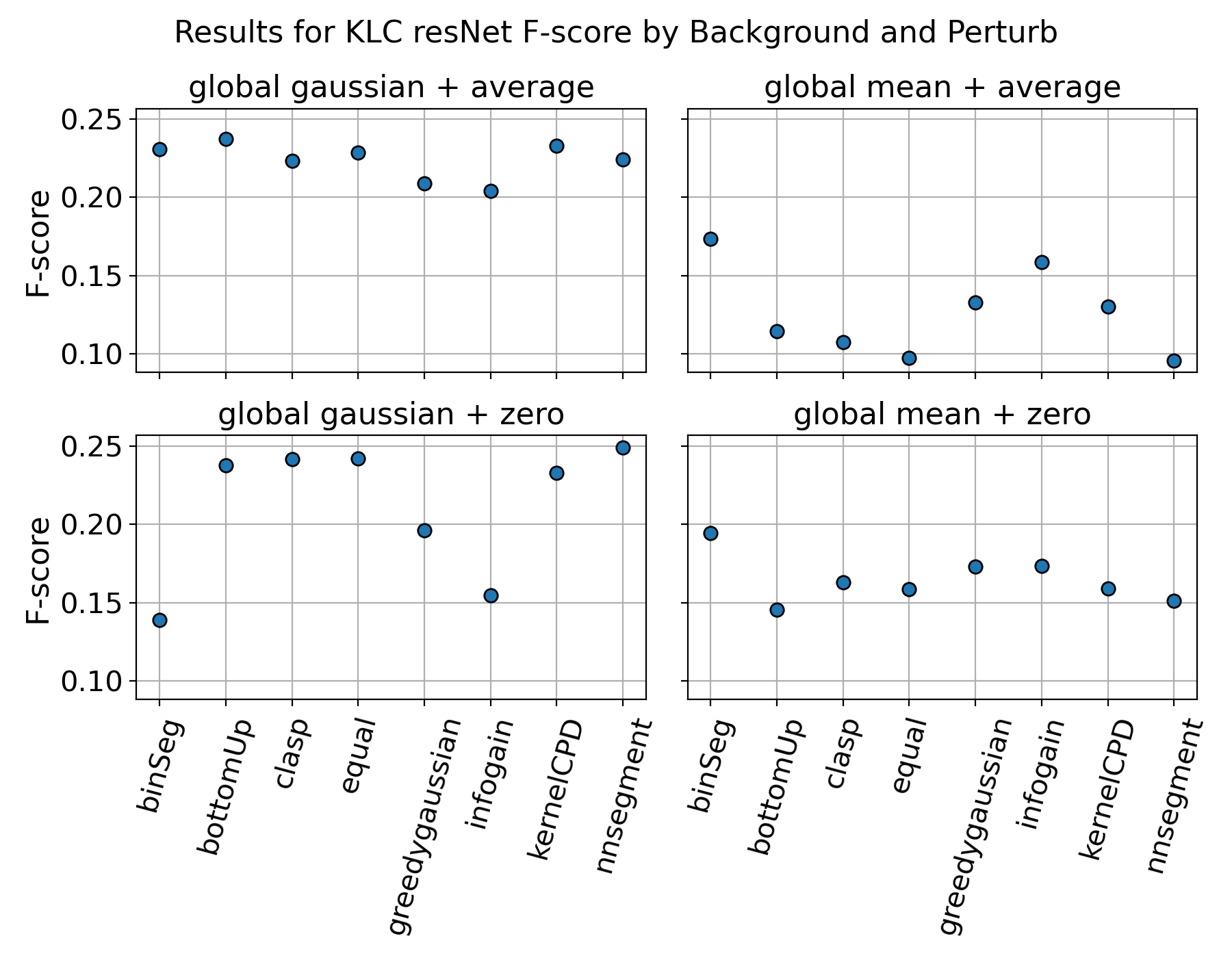}
    \caption{F-score on ResNet and KeplerLightCurves by Background and Perturbation over Segmentations. Different Segmentations perform better with different backgrounds and perturb methods}
    \label{fig:KPLRES1}
\end{figure}

Figure \ref{fig:KPLRES1} shows the F-score results for the ResNet classifier on the KLC dataset for different segmentations, backgrounds and perturbation methods. Overall, global Gaussian perturbations generally produced higher results compared to global mean. The choice between average and zero background also had a significant impact on performance. When using global mean perturbations, the average background consistently led to lower scores across most segmentation methods. However, this trend reversed under global Gaussian perturbations, where zero background often yielded higher results. This suggests that zero background synergises better with global Gaussian perturbations, while average background may be more compatible with global mean.

The perturbation type played a critical role in determining segmentation performance. Global Gaussian perturbations, which introduce noise, generally led to higher scores, implying better performance for most methods. In contrast, global mean perturbations, which adjust the mean value, resulted in lower scores across the board, indicating reduced effectiveness. This variability suggests that method performance is highly dependent on the combination of background and perturbation type.

Comparing the different segmentations, we see that the Binary, Infogain, and Greedy Gaussian segmentations underperform compared to the other methods when the global Gaussian + zero background strategy is used. However, these three segmentations outperform others when the global mean is used, especially with the average background. This suggests that the optimal choice for segmentation can be dependent on the perturbation strategy used in evaluation. 

\begin{figure}
    \centering
    \includegraphics[width=1.0\linewidth]{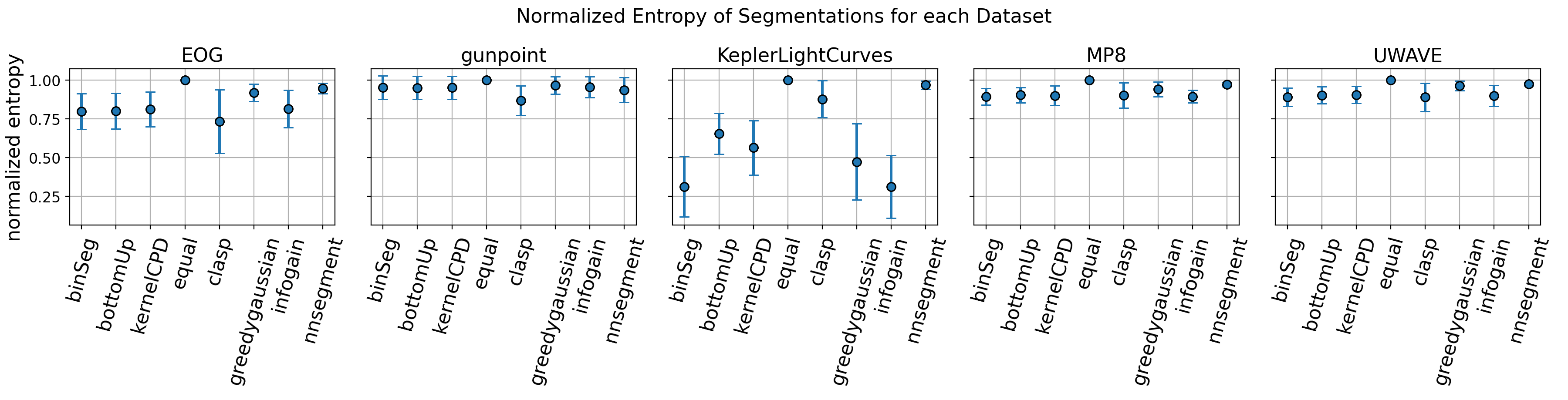}
    \caption{Mean normalised entropy by Dataset and Segmentation. Error bar denotes standard deviation. Segmentations significantly deviate from uniform in KeplerLightCurves}
    \label{fig:entropy}
\end{figure}

To further understand what makes these segmentations behave differently on KLC, we calculated the normalised entropy of each segmentation, which is based on the segment length compared to the equal segmentation as a baseline. Figure \ref{fig:entropy} shows the entropy results across all segmentations and datasets. We can see that for KLC, the same three segmentations have a lower score, indicating a greater deviation from the equal segmentation. This means that these segmentation methods tend to find a highly uneven distribution of segment lengths, including very large and very small segments.  As segments become smaller, each of their means can significantly deviate from the global mean, which influences the result of the perturbations based on the global mean.

\subsubsection{Background.}
\begin{figure}[h!]
    \centering
    \begin{subfigure}{0.49\textwidth} 
        \includegraphics[width=1\linewidth]{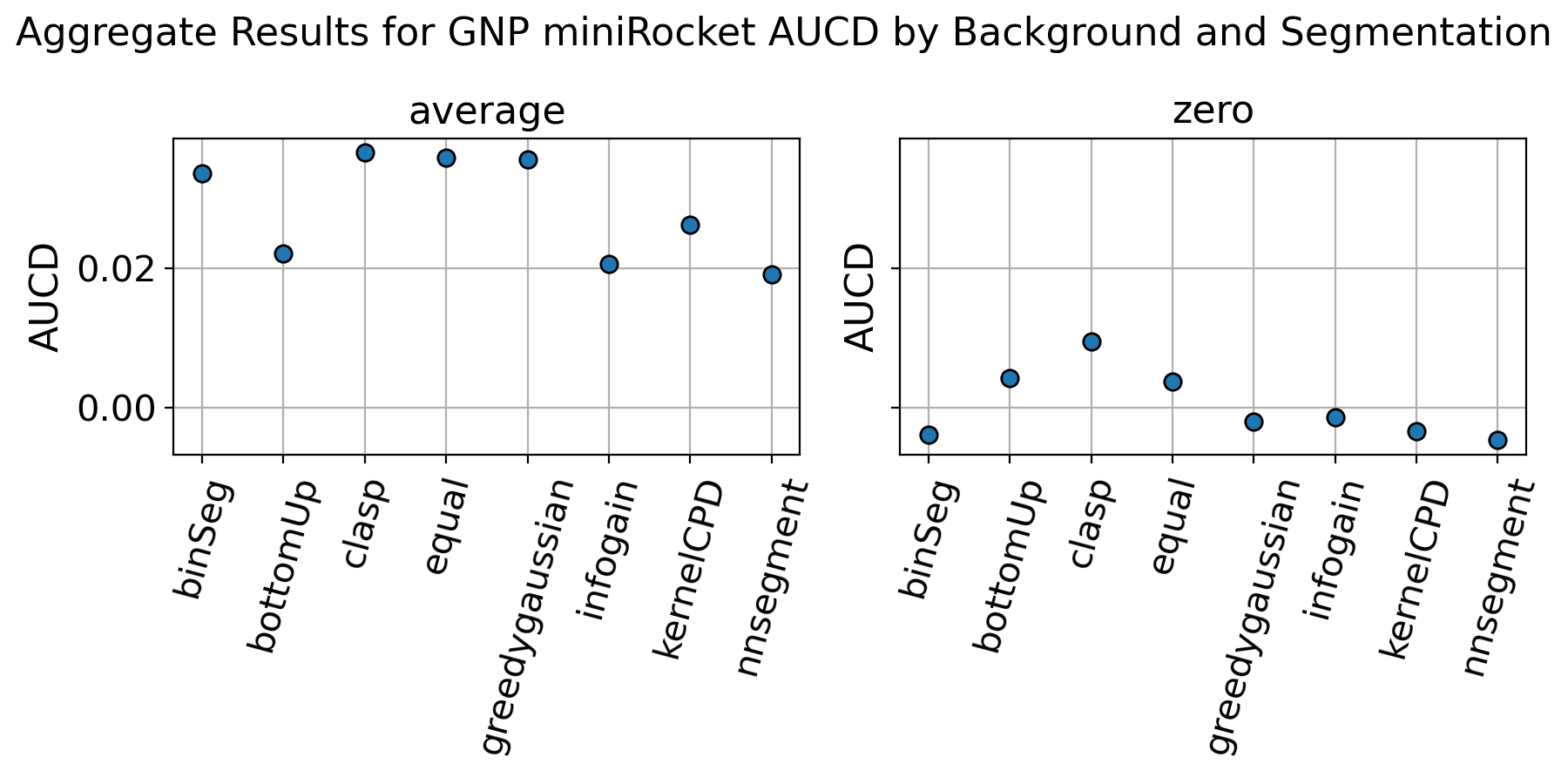} 
        \caption{AUCD for MiniRocket on Gunpoint over Background \& Segmentation}
        \label{fig:AUCD_miniRcket_gunpoint}
    \end{subfigure}
    \begin{subfigure}{0.49\textwidth}
        \includegraphics[width=1\linewidth]{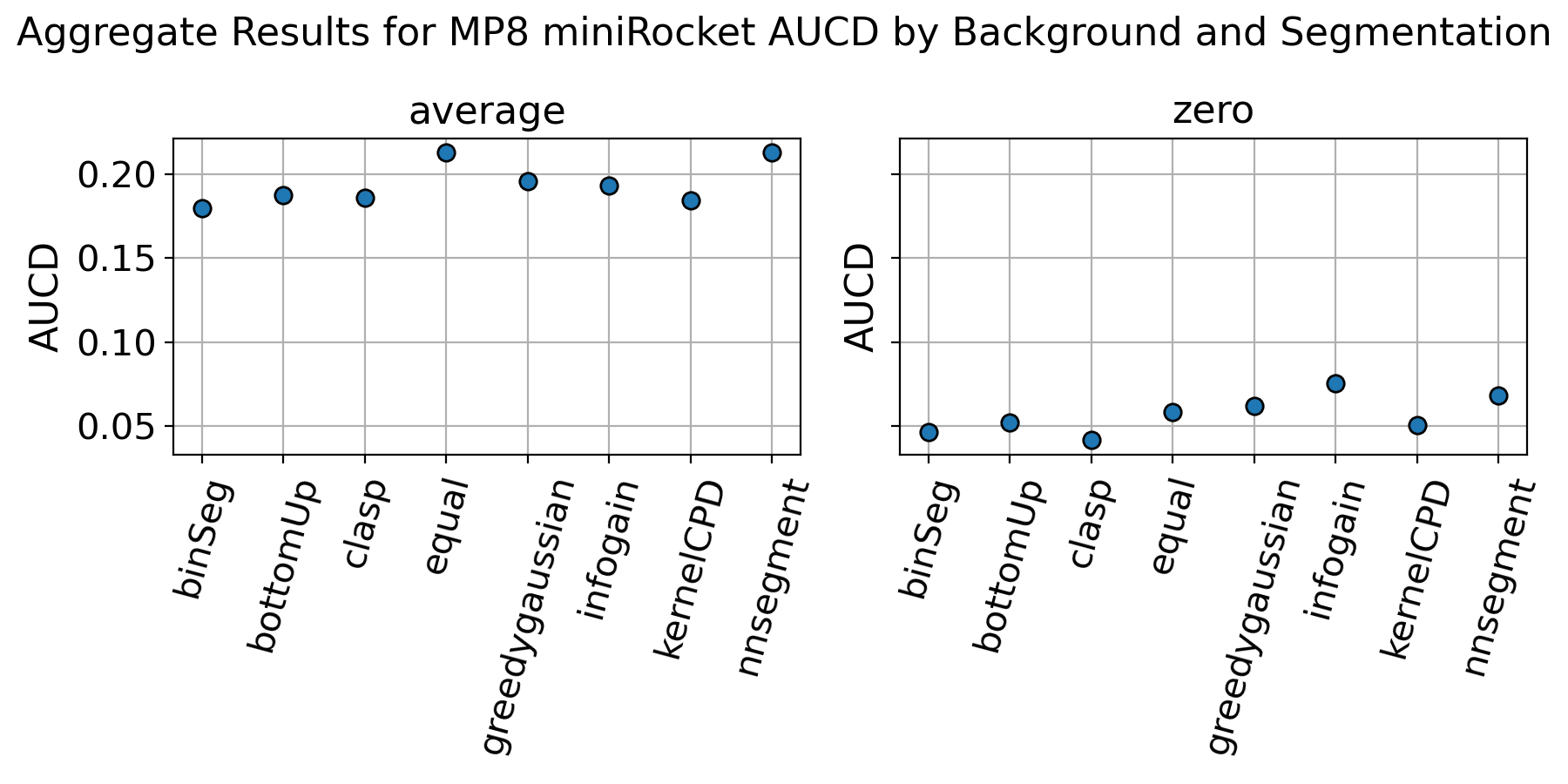} 
        \caption{AUCD for MiniRocket on MilitaryPress over Background \& Segmentation}
        \label{fig:AUCD_MP_miniRocket}
    \end{subfigure}
    \caption{AUCD for MiniRocket on Gunpoint and MilitaryPress. Both cases show that the average background outperforms the zero background.}
    \label{fig:background}
\end{figure}

Figures \ref{fig:AUCD_miniRcket_gunpoint} and \ref{fig:AUCD_MP_miniRocket} respectively show the AUCD results for MiniRocket on the GunPoint and MilitaryPress datasets. 
The results show that using the average background over the zero background yields better results for AUCD across all Segmentations. The advantage of the average background is that it fundamentally provides a more realistic baseline that reflects the typical behaviour of the data. A zero background assumes all features are absent or neutral, which can lead to misleading interpretations if the actual data never contains such values. In contrast, the average sample captures the distribution of the data, ensuring SHAP values explain predictions relative to a meaningful reference point.

\subsection{Attribution Normalisation}

\begin{figure}[h]
    \centering
    \includegraphics[width=1\linewidth]{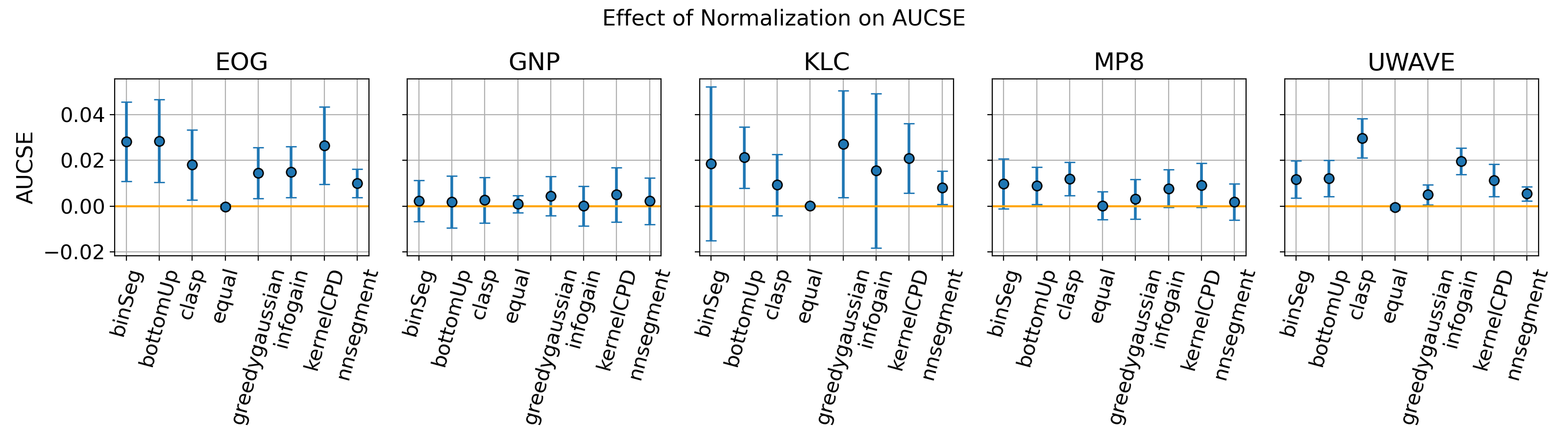} 
    \caption{Difference in AUCSE results when normalization is applied. A positive value denotes an increase in mean evaluation performance with normalization. Error bars are standard deviation of the difference.}
    \label{fig:normalization}
\end{figure}

Figure \ref{fig:normalization} shows that \emph{normalisation} appears to be particularly beneficial across the board, with notable performance gains for the ClaSP, Binary, Greedy Gaussian and Infogain  Segmentations. Normalisation's preservation of segment-level additivity appears to have a clear positive effect for most datasets, ensuring timepoint attributions sum to the segment’s SHAP value. This prevents distortions in importance rankings, especially for varying segment lengths, leading to more reliable and interpretable explanations. 
Because the normalisation is fundamentally based on compensating based on segment length, as the partition of a time series into segments becomes more uniform (equal-sized segments), its effect in changing the relative rankings of points is more muted. And since the evaluation measures are based on the relative scale of attributions,  the closer a segmentation is to equal, the more invariant the resulting evaluation is to the normalisation process. We can observe this empirically by seeing that the equal segmentation yields practically no change in evaluation between normalised and unnormalised explanations. 

We can measure how close a segmentation is to the uniform segmentation through its relative entropy. We show the expected relative entropy for each segmentation and dataset in Figure \ref{fig:entropy}. We can see that the datasets and segmentations that gain the most from normalisation, EOG and KeplerLightCurves, are also the ones whose relative entropy is most different from the uniform. This indicates that normalisation is especially effective when there is a mix of uneven segmentation, and when the segmentation is even, normalisation leaves evaluation unchanged, making it a desirable and inexpensive way to improve explanations on segmentations.

\subsection{Differences between XAI Evaluation Measures}

\begin{figure}[h!] 
    \centering
    \begin{subfigure}[b]{0.49\textwidth} 
        \includegraphics[width=\linewidth]{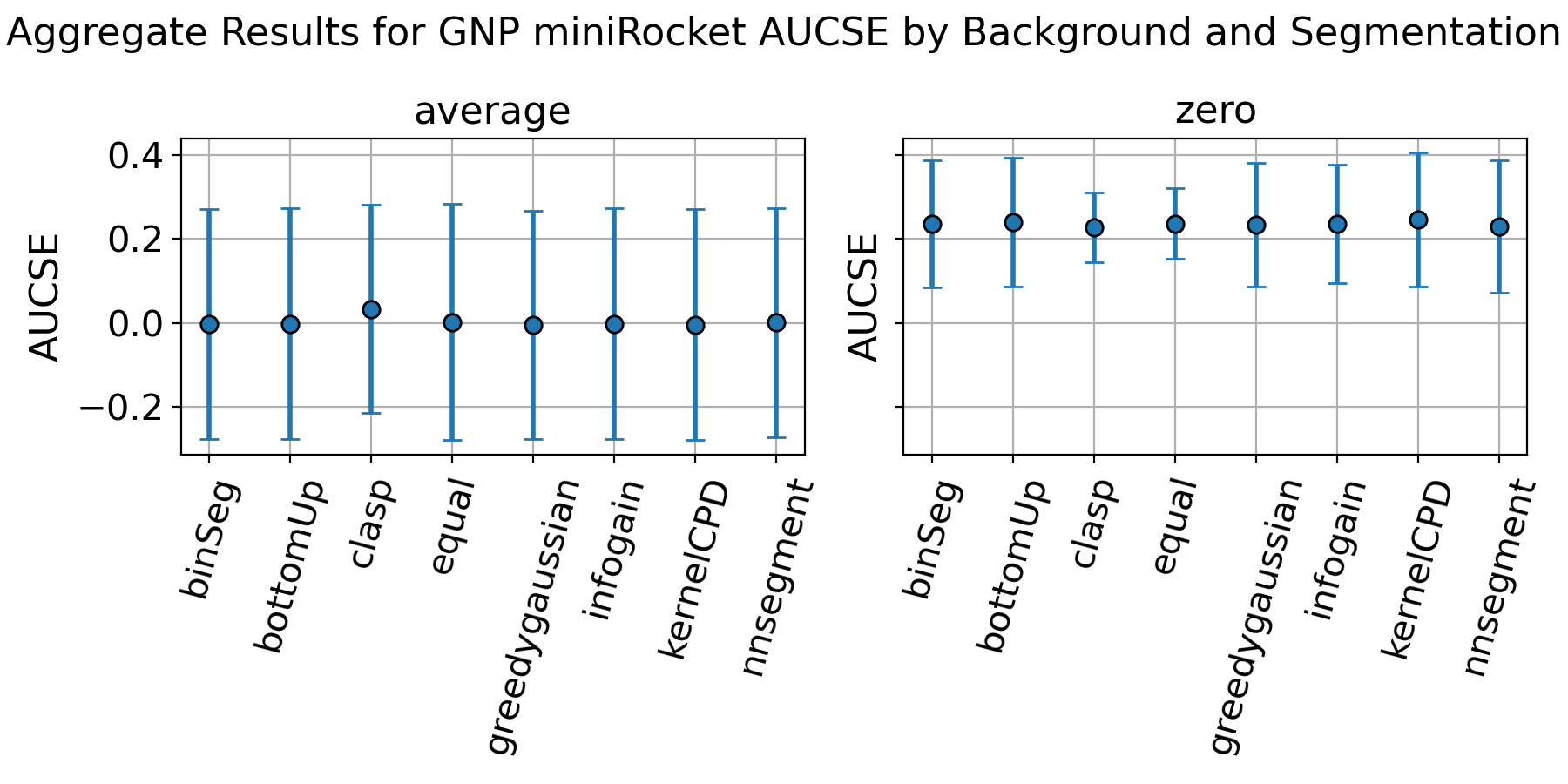} 
        \caption{AUCSE results for MiniRocket on Gunpoint for different backgrounds and segmentations. Zero background outperforms the average background.}
        \label{fig:backgroundScores4_gunpoint_miniRocket}
    \end{subfigure}
    \begin{subfigure}[b]{0.49\textwidth}
        \includegraphics[width=\linewidth]{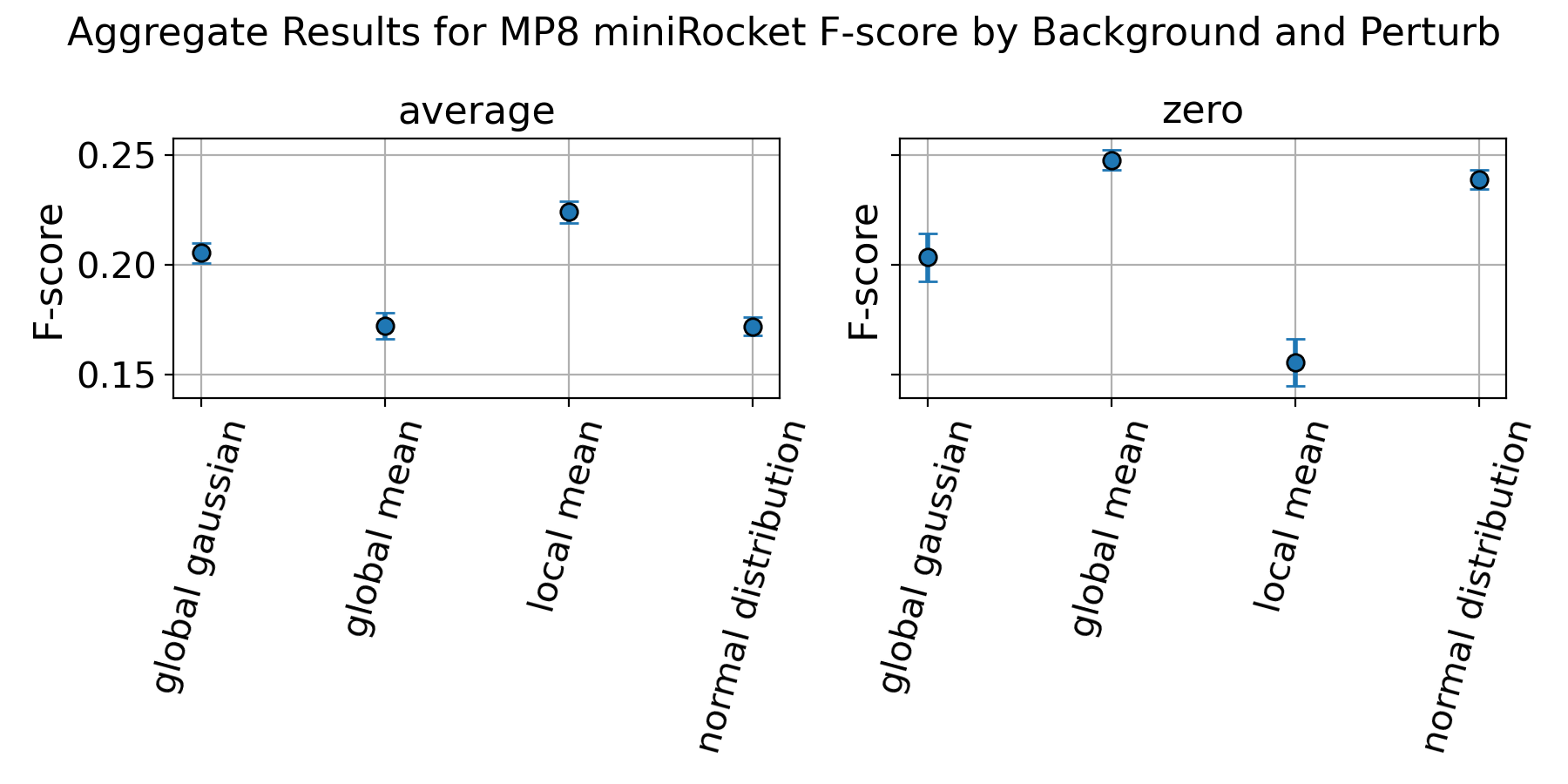} 
        \caption{F-score results for MiniRocket on MP8 for different backgrounds and perturbations. Best choice of background depends on perturbation choice in evaluation.}
       \label{fig:MP_minirocket_AUCSE}
    \end{subfigure}
    \caption{InterpretTime's scores for Gunpoint MiniRocket and for MP MiniRocket.}
\end{figure}

\begin{figure}[h!] 
    \centering
    \begin{subfigure}[b]{0.35\textwidth} 
        \includegraphics[width=\linewidth]{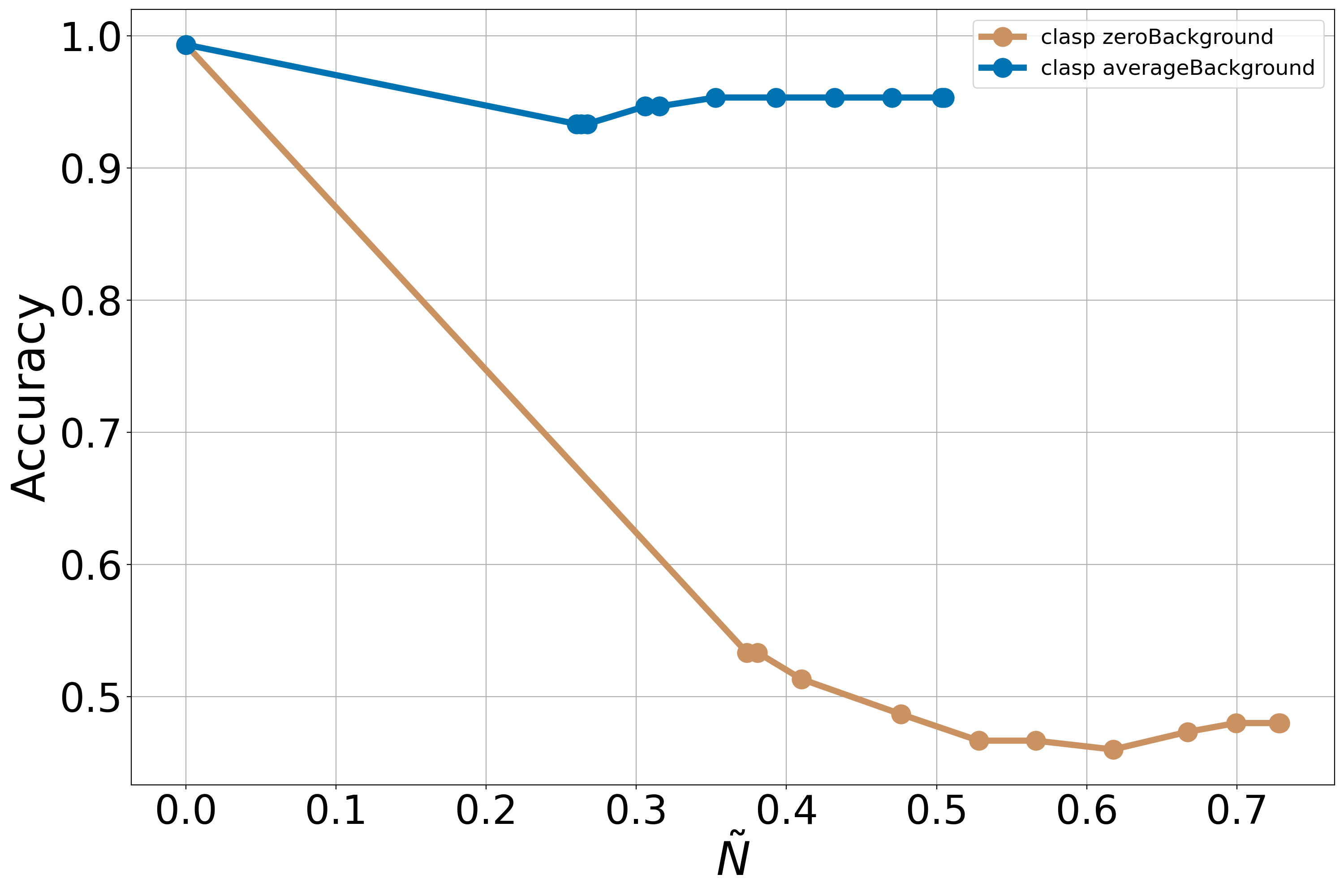}
        \caption{Accuracy decay while corrupting time series}
        \label{fig:interpret_plots_miniRocket_gunpoint_accuracy_decrease}
    \end{subfigure}
    \begin{subfigure}[b]{0.59\textwidth}
        \includegraphics[width=\linewidth]{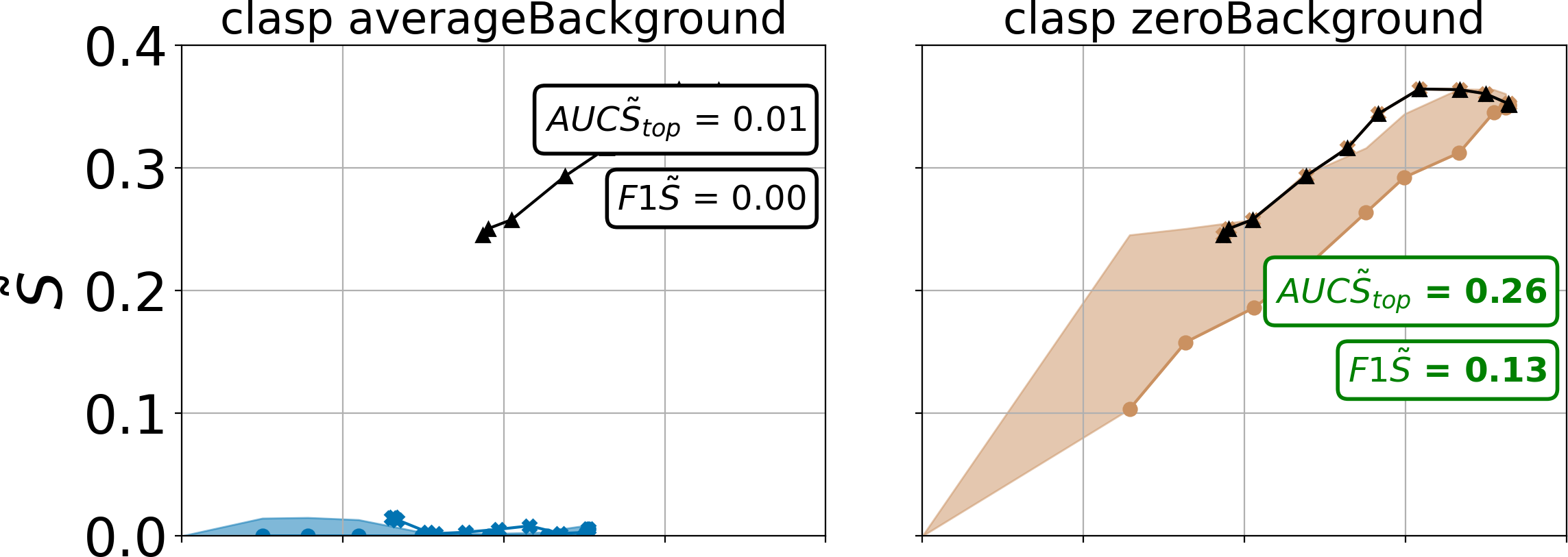}
        \caption{Evolution of $\bar{S}$ while corrupting TS.}
        \label{fig:interpret_plots_miniRocket_gunpoint_delta}
    \end{subfigure}
    \caption{Accuracy decay and probability of the initial predicted label evolutions while the time series are progressively corrupted for both backgrounds, ClaSP segmentation, MiniRocket classifier, Gunpoint dataset. }
    \label{fig:interpret_plots_miniRocket_gunpoint}
\end{figure}

In this section, we explore the few cases where the evaluation measures substantially disagree. While AUCSE and F-score agree in all analysed cases, there are some differences between these two and AUCD, which we are investigating in this section. 

\textbf{Gunpoint MiniRocket}: 
As can be noted from Figure \ref{fig:backgroundScores4_gunpoint_miniRocket}, according to InterpreTime evaluation measure, the zero background is the best choice, different from what was previously observed for AUCD in Figure \ref{fig:AUCD_miniRcket_gunpoint}. \\
We further investigate the evolution using Figure \ref{fig:interpret_plots_miniRocket_gunpoint_accuracy_decrease}, showing the accuracy and Figure \ref{fig:interpret_plots_miniRocket_gunpoint_delta} showing the probability of the predicted label evolution. 
This analysis tracks how InterpretTime progressively corrupts the time series for two different instances that share the same ClaSP segmentation but use different background sets.
The first plot reveals a violation of InterpretTime’s core assumption: the instance using the average background after a slight initial accuracy decay rises again, never reaching the random accuracy level of the zero background. This directly impacts the evaluation measure, as evidenced by the flat curve in the second plot. Since AUCSE is defined as the evolution of $\bar{S}$ through the pre-defined steps, if this evolution is marginal or absent, the AUCSE measure would be close to 0.

\textbf{MP8 MiniRocket}: Figure \ref{fig:MP_minirocket_AUCSE} shows the F-score (AUCSE is very similar) grouped by different perturbations. 
The zero background achieves higher scores using the global mean and normal distribution perturbations, whereas using the global Gaussian and local mean yields the opposite result. At the same time, AUCD constantly scores the average background as the best choice as evident from Figure \ref{fig:AUCD_MP_miniRocket}.  \\
Since the majority of the analysed evaluation measures agree that the average is the best choice, we can conclude that this is one example where different perturbations yield different results.

\begin{figure}[t] 
    \centering

    \begin{subfigure}[b]{0.49\textwidth}
        \includegraphics[width=\linewidth]{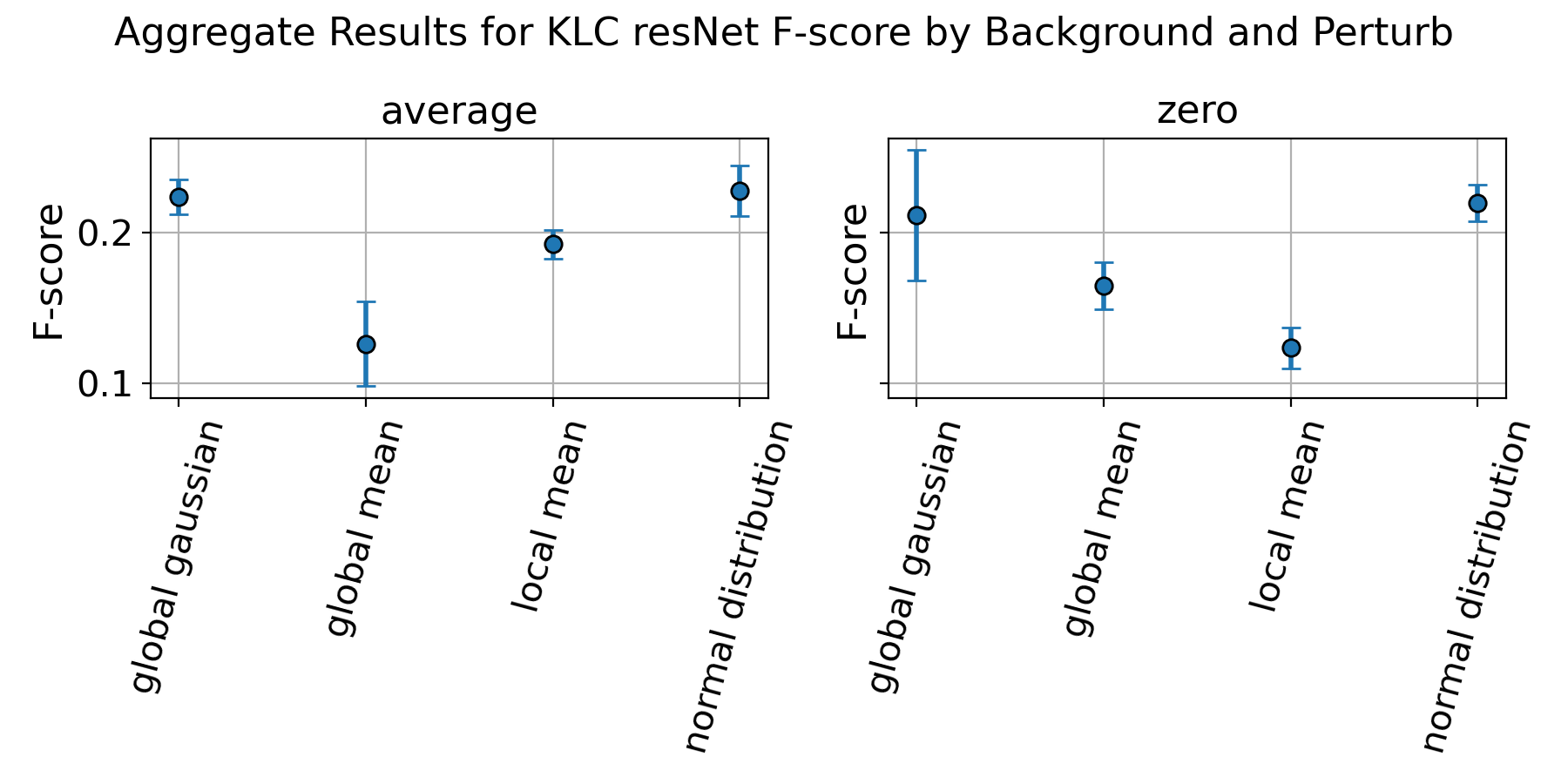} 
        \caption{ResNet to KLC: F-score of both backgrounds for each used perturbation}
        \label{fig:KLC_ResNet_F1}
    \end{subfigure}
    \begin{subfigure}[b]{0.49\textwidth} 
        \includegraphics[width=\linewidth]{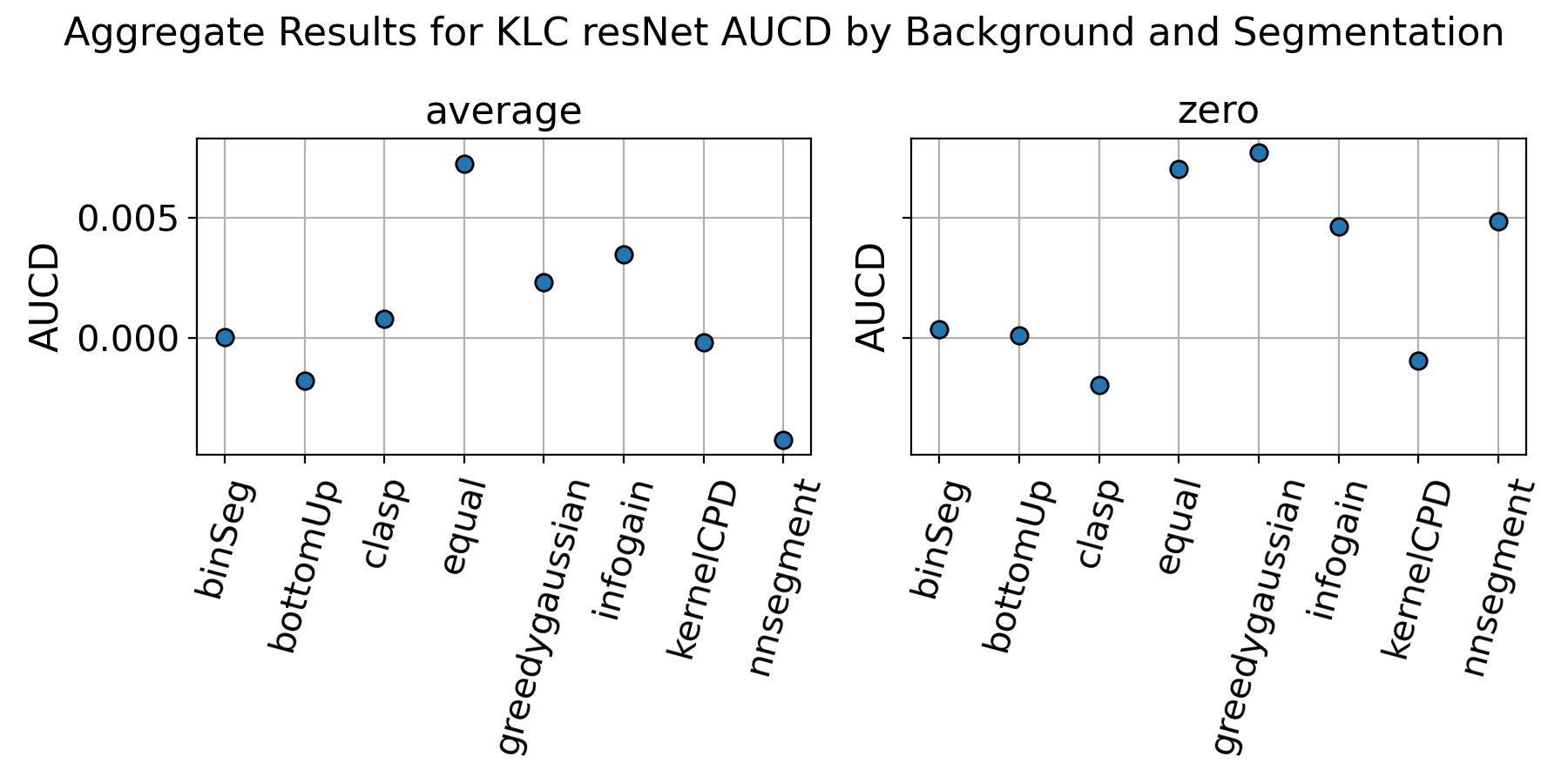} 
        \caption{ResNet applied to KLC AUCD, backgrounds and segmentations}
        \label{fig:KLC_ResNet_AUC-diff}
    \end{subfigure}

    \caption{InterpretTime's and AUCD scores for (a) and (b) ResNet applied to KLC, (c) and (d) MiniRocket applied to MP} 
    \label{fig:MP_gunpoint_differentPerturb}

\end{figure}

\textbf{KLC ResNet}:  
Finally we investigate the third combination in this section,
i.e., ResNet classifier and KLC dataset. In Figure \ref{fig:KLC_ResNet_F1}  the F-score is reported (AUCSE has similar behavior, having a smaller difference) grouped by different perturbations, while Figure \ref{fig:KLC_ResNet_AUC-diff} shows AUCD, grouped by segmentation algorithms. 
Using these last evaluation measure, the different segmentation methods achieve, most of the time, very similar results, with the exceptions of Greedy Gaussian and NNSegment having slightly higher scores when the zero background is used. \\
Looking instead at $F$-scores, there is a completely different scenario: using global Gaussian and normal distribution perturbations, the two backgrounds achieve nearly the same score; on the other hand, global mean and local mean indicate two opposite best choices, respectively, zero and average background. 

Our findings suggest that noise augmentation during training, as proposed in the original InterpretTime paper, plays a key role in ensuring consistent rankings (see \cite{turbe2023evaluation} and \cite{seramazza24} for details). Consequently, this limits the scope of this method to neural networks whose performance is not hurt by the noise addition. 

\section{Conclusion}
In this work, we analysed key factors influencing the explanation resulting from SHAP in the Time Series domain. Our analysis was conducted using a diverse set of classifiers, datasets, backgrounds, and segmentation methods. 

We empirically showed that the number of segments is more important than the underlying segmentation method used.  
Among the eight analysed segmentation methods, the clear winner is the equal-length segmentation, showing superior or similar results to the other methods. This is mostly achieved because the other methods often produce very similar segmentations, except for the KLC dataset, where some segmentations are very uneven, while requiring more computational resources. 

Another important contribution is the \textit{attribution normalisation} technique, which consists of weighting each segment contribution according to its size, which increases the XAI evaluation scores in the case of uneven segmentations. 

We also found small differences between the results achieved using the zero and the average background. In the few cases we found substantial differences between their relative scores, the average background was the best choice. \\
Moreover, in preliminary experiments, we also tested the \textit{sampling background} strategy, consisting of a set composed of different random samples from the train set: we found a marginal improvement over the other two backgrounds, but at the cost of much higher computational cost. An interesting direction for future work is to further study  the effect of the background set used by SHAP, an important topic still under-explored in the literature. 

Finally, we observe some inconsistency between the two XAI evaluation methods, suggesting that InterpretTime’s effectiveness may be limited to deep learning classifiers. Unlike other models, deep learning classifiers remain unaffected by the required noise-based training augmentation. Without this augmentation, i.e., as used in this work, results may vary unpredictably across different perturbations.

\section*{Acknowledgements} 
This publication has emanated from research
conducted with the financial support of Taighde
Éireann – Research Ireland under Grants [ML-Labs 18/CRT/6183, Insight Centre for Data Analytics 12/RC/2289\_P2]. For the purpose of Open Access, the author has applied a CC BY
public copyright license to any Author Accepted Manuscript version arising from this
submission.
%
%
%
%


\bibliographystyle{splncs04}
\bibliography{bib}

\end{document}